\def\year{2022}\relax
\documentclass[letterpaper]{article} 
\usepackage{aaai22}  
\usepackage{times}  
\usepackage{helvet}  
\usepackage{courier}  
\usepackage[hyphens]{url}  
\usepackage{graphicx} 
\usepackage{natbib}  
\usepackage{caption} 
\DeclareCaptionStyle{ruled}{labelfont=normalfont,labelsep=colon,strut=off} 
\usepackage[ruled,linesnumbered,noend]{algorithm2e}
\usepackage{multirow}
\usepackage{subcaption}
\usepackage{amsmath,amssymb,amsthm}
\usepackage{bbm}
\usepackage{subfiles}
\newtheorem{lemme}{Lemma}

\title{An Accurate HDDL Domain Learning Algorithm from Partial and Noisy Observations}
\author{
 Maxence Grand,
 Damien Pellier,
 Humbert Fiorino\\
 }
 \affiliations {
 Univ. Grenoble Alpes, LIG, Grenoble, France\\
 \{Maxence.Grand, Damien.Pellier, Humbert.Fiorino\}@univ-grenoble-alpes.fr
 }

\begin{document}
\maketitle

\begin{abstract}
    The Hierarchical Task Network ({\sf HTN}) formalism is very expressive and used to express a wide variety of planning problems. In contrast to the classical {\sf STRIPS} formalism in which only the action model needs to be specified, the {\sf HTN} formalism requires to specify, in addition, the tasks of the problem and their decomposition into subtasks, called {\sf HTN} methods. For this reason, hand-encoding {\sf HTN} problems is considered more difficult and more error-prone by experts than classical planning problem. To tackle this problem, we propose a new approach (HierAMLSI) based on grammar induction to acquire {\sf HTN} planning domain knowledge, by learning action models and {\sf HTN} methods with their preconditions. Unlike other approaches, HierAMLSI is able to learn both actions and methods with noisy and partial inputs observation with a high level or accuracy.
\end{abstract}

\section{Introduction}

The Hierarchical Task Network ({\sf HTN}) formalism \cite{DBLP:conf/aaai/ErolHN94} is very expressive and used to express a wide variety of planning problems. This formalism allows planners to exploit domain knowledge to solve problems more efficiently \cite{DBLP:journals/expert/NauAIKMMWY05} when planning problems can be naturally decomposed hierarchically in terms of tasks and task decompositions. The standard language used to model {\sf HTN} problem is HDDL (Hierachical Domain Description Language) \cite{hddl}. In contrast to the classical PDDL language used to model {\sf STRIPS} problems in which only the action model needs to be specified, HDDL requires to specify the task model of the problem. A task model can be primitive and compound. A primitive task model is described by PDDL operators. A compound tasks model is described using {\sf HTN} methods. An {\sf HTN} method describes the set of primitive and/or compound task required to decompose a specific compound task. For this reason, hand-encoding {\sf HTN} problems is considered more difficult and more error-prone by experts than classical planning problem. This makes it all the more necessary to develop techniques to learn {\sf HTN} domains.

Many machine learning approaches have been proposed to facilitate the acquisition of PDDL domain acquisition and to learn the underlying action model, e.g, ARMS \cite{arms}, FAMA \cite{fama}, LOCM \cite{locm}, LSONIO \cite{lsonio}, AMLSI \cite{amlsi_keps, amlsi_ictai}. In these approaches, training data are either (possibly noisy and partial) intermediate states and plans previously generated by a planner, or randomly generated action sequences (i.e. random walks). On the other hand, few approaches have been proposed to learn {\sf HTN} domains. However, it is possible to mention  CAMEL \cite{camel}, HTN-Maker \cite{htn-maker, htn_maker_nd}, LHTNDT \cite{LHTNDT} or HTN-Learner \cite{htn_learner}. The major drawbacks of these approaches are: (1) they consider to have complete and noiseless observations as input; (2) they only learn {\sf HTN} methods except HTN-Learner, i.e., they consider that the action model is known a priori and (3) the learned domains are not {\em accurate} enough to be used "as is" in a planner. A step of expert proofreading is still necessary to correct them. Even small syntactical errors can make sometime the learned domains useless for planning

To deal with these drawbacks, we propose in this paper, a new learning algorithm for HDDL domains, called HierAMLSI. HierAMLSI is based on AMLSI \cite{amlsi_keps, amlsi_ictai}, a PDDL domain learner based on grammar induction. HierAMLSI takes as input a set of partial and noisy observations and learns a full HDDL planning domain with action model and {\sf HTN} methods. We show experimentally that HierAMLSI is highly accurate even with highly partial and noisy learning datasets minimising HDDL domain proofreading by experts. In many HDDL ICP benchmarks HierAMLSI does not require any correction of the learned domains at all.





The rest of the paper is organized as follows. In section \ref{sec:formal} we present the problem statement. In section \ref{sec:backgroung} we give some backgrounds on the AMLSI approach. In section \ref{sec:method}, we detail the HierAMLSI steps. Then, section \ref{sec:experiments} evaluates the performance of HierAMLSI on IPC benchmarks. Finally, Section \ref{sec:related} concludes with the related works.

\section{Formal Framework}\label{sec:formal}

Section \ref{sec:formal:strips} introduces a formalization of {\sf STRIPS} planning domain learning consisting in learning a transition function of a grounded planning domain and in expressing it as PDDL operators and Section \ref{sec:formal:htn} extends this formalization to {\sf HTN} domains.

\subsection{STRIPS Planning}\label{sec:formal:strips}

In this section we use definitions and notations proposed by \cite{holler16} and adapt them to the learning problem.

A {\sf STRIPS} planning problem is a tuple $P = (L, A, S, s_0, s, \delta, \lambda)$, where $L$ is a set of logical propositions describing the world states, $S$ is a set of state labels, $s_0 \in S$ is the label of the initial state, and $g \subseteq S$ is the set of goal labels. $\lambda$ is an observation function $\lambda : S \rightarrow 2 ^L$ that assigns to each state label the set of logical propositions true in that state. $A$ is a set of action labels.Action preconditions, positive and negative effects are given by the functions $prec$, $add$ and $del$ that are included in $\delta = (prec, add, del)$.  $prec$ is defined as $prec : A \rightarrow 2^L$. The functions $add$ and $del$ are defined in the same way. Without loss of generality, we chose this unusual formal framework inspired by \cite{holler16} in order to define the {\sf STRIPS} learning problem as the lifting of a state transition system into a propositional language.

The function $\tau : A \times S \rightarrow \{true , false \}$ returns whether an action is applicable to a state, i.e. $\tau (a, s) \Leftrightarrow prec(a) \subseteq \lambda(s)$. Whenever action $a$ is applicable in state $s_i$, the state transition function $\gamma : A \times S \rightarrow S$ returns the resulting state $s_{i+1}=\gamma(s_i,a)$ such that $\lambda(s_{i+1}) = [\lambda(s_i) \setminus del(a)] \cup add(a)$.

A sequence $(a_0 a_1 \dots a_n )$ of actions is applicable to a state $s_0$ when each action $a_i$ with $0 \leq i \leq n$ is applicable to the state $s_i$. Given an applicable sequence $(a_0 a_1 \dots a_n )$ in state $s_0$, $\gamma(s_0, (a_0 a_1 \dots a_n )) = \gamma(\gamma(s_0, a_0), (a_1 \dots a_n ))= s_{n+1}$. It is important to note that this recursive definition of $\gamma$ entails the generation of a sequence of states $(s_0 s_1 \dots s_{n+1} )$. A goal state is a state $s$ such that $g \in G$ and $\lambda(g) \subseteq \lambda(s)$. $s$ satisfies $g$, i.e. $s \models g$ if and only if $s$ is a goal state. An action sequence is a solution plan to a planning problem $P$ if and only if it is applicable to $s_0$ and entails a goal state.

In formal languages, a set of rules is given that describe the structure of valid words and the language is the set of these words. For {\sf STRIPS} planning problem $P = (L, A, S, s_0, G, \delta, \lambda)$, this language is defined as ($0 \leq i \leq n$):
\begin{equation*}
    {\cal L}(P) = \{\omega = (a_0a_1 \dots a_n )| a_i \in A, \gamma(s_0, \omega) \models g\}
\end{equation*}

We know that the set of languages generated by {\sf STRIPS} planning problems are regular languages \cite{holler16}. In other words, a {\sf STRIPS} planning problem $P = (L, A, S, s_0, G, \delta, \lambda)$ generates a language ${\cal L}(P)$ that is equivalent to a Deterministic Finite Automaton (DFA) $\Sigma = (S, A, \gamma)$. $S$ and $A$ are respectively the nodes and the arcs of the DFA, and $\gamma$ is the transition function.

\begin{figure}
    \begin{center}
    \includegraphics[scale=1]{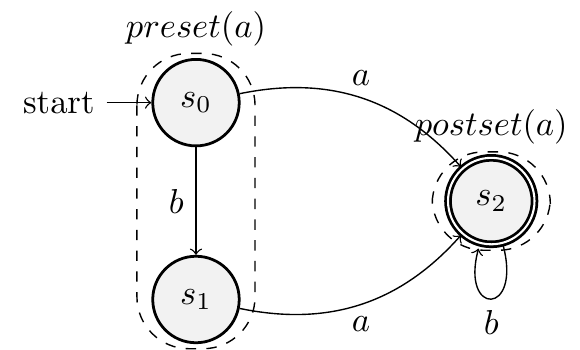}
    \end{center}
    \caption{An example of DFA with pre-states and post-states}
    \label{fig:dfa}
\end{figure}

For any arc $a \in A$, we call {\em pre-set} of $a$ the set $preset(a) = \{s \in S \ | \ \gamma(s, a)=s' \}$ and {\em post-set} of $a$ the set $postset(a) = \{s' \in S \ | \ \gamma(s, a)=s'\}$ (see Figure  \ref{fig:dfa}).

A {\sf STRIPS} learning problem is as follow: given a set of observations $\Omega \subseteq {\cal L}(P)$, is it possible to learn the DFA $\Sigma$, and then infer $P$?

For instance, suppose $\Omega = \{a,ab, ba, bab, abb,\dots\}$ such that
$s_0 \xrightarrow{\text{a}} s_2$,
$s_0 \xrightarrow{\text{a}} s_2 \xrightarrow{\text{b}} s_2$,
$s_0 \xrightarrow{\text{b}} s_1 \xrightarrow{\text{a}} s_2$,
$s_0 \xrightarrow{\text{b}} s_1 \xrightarrow{\text{a}} s_2 \xrightarrow{\text{b}} s_2$,
$s_0 \xrightarrow{\text{a}} s_2 \xrightarrow{\text{b}} s_2 \xrightarrow{\text{b}} s_2 \ldots$
\cite{amlsi_ictai} show that it is possible to learn $\Sigma$ (see~Figure~\ref{fig:dfa}) and infer $P$ with actions $\{a,b\}$, the initial state $s_0$ and some states marked as goal $G = \{s_2\}$.

\subsection{HTN Planning}\label{sec:formal:htn}



By extension based on the notation of \cite{holler21}, an {\sf HTN} planning problem is a tuple $P = (L, C, A, S, M, s_0, w_I, g, \delta, \tau, \lambda, \sigma, \zeta)$. As for {\sf STRIPS} problems, $L$ is a set of logical propositions describing the world states, $S$ is a set of state labels, $s_0 \in S$ is the label of the initial state, $g \subseteq S$ is the set of goal label, $\lambda$ is the observation function and preconditions, positive and negative effects are given by the functions $prec$, $add$ and $del$ included in $\delta$.

$A$ is the set action (or primitive task) labels and $C$ is a set of compound (or non primitive) task labels, with $C \cap A = \emptyset$.  Tasks are maintained in {\it task networks}. A task network is a sequence of tasks (for simplicity, we consider only Totally Ordered domain). Let $T = C \cup A$. A task network is an element out of $T^{*}$ ($^*$ is the Kleene operator). Compound tasks are decomposed using methods. The set $M$ contains all method labels. Methods are defined by the function $\sigma : M \rightarrow C \times T^{*}$. Then, a coumpound task $c$ is decomposable in a state $s$ if and only if there exists a revelant method $m \in M$ such that: $\sigma(m) = (c,\phi)$ and $prec(m) \in s$. The function $\zeta : T^{*} \times S \rightarrow T^{*}$ gives the decomposition function. For a totally orderer task network $\omega = \omega_1 t \omega_2$, $\zeta$ is defined as follows:
\begin{equation*}
\resizebox{\hsize}{!}{$
\begin{array}{l}
     \zeta(\omega_1 t \omega_2, s) = \left\{
    \begin{array}{ll}
         \omega_1 t \omega_2 & \mbox{if $t$ is a primitive task}\\
         \omega_1 \phi \omega_2 & \mbox{if $t$ is a coumpound task}\\
         & \mbox{and $t$ is decomposable in $\gamma(\omega_1,s)$}\\
         \emptyset & \mbox{Otherwise}
    \end{array}\right.
\end{array}$}
\end{equation*}
As $\omega_1 t \omega_2$ is a totally ordered task network, $\omega_1$ contains only primitive tasks. We denote $\omega \rightarrow^{*} \omega'$ that $\omega$ can be decomposed into $\omega'$ by 0 or more method applications. Finally, $\omega_I$ is the initial task network.

A solution to an {\sf HTN} planning problem is a task network $\omega$ with:
\begin{enumerate}
    \item $\omega_I \rightarrow^{*} \omega$, i.e. it can be reached by decomposing $\omega_I$.
    \item $\omega \in A^{*}$, i.e. all tasks are primitive.
    \item $\gamma(s_0,\omega) \models g$, i.e. $\omega$ is applicable in $s_o$ and results in a goal state.
\end{enumerate}

Finally, we can define an {\sf HTN} planning problem $P = (L, C, A, S, M, s_0, w_I, g, \delta, \tau, \lambda, \sigma, \zeta)$ as a formal language:
\begin{equation*}
    \resizebox{\linewidth}{!}{$
    \begin{array}{l}
{\cal L}(P) = \{\omega = (t_0t_1 \dots t_n)| t_i \in A, \gamma(s_0, \omega) \models g, \omega_I \rightarrow^{*} \omega\}\\
    \end{array}
    $}
\end{equation*}


An {\sf HTN} learning problem is as follow: given a set of observations $\Omega \subseteq {\cal L}(P)$, is it possible to learn $P$?

Unlike {\sf STRIPS} planning problems, the language ${\cal L}(P)$ is not necessary regular \cite{holler14} and cannot be represented as a DFA. As mentioned by \cite{holler14, holler21}, ${\cal L}(P)$ is the intersection of two languages:
\begin{enumerate}
    \item ${\cal L_H}(P) = \{\omega \in A^* | w_I \rightarrow^* \omega\}$ which is defined by the decomposition hierarchy, i.e. by the compound tasks and methods.
    \item ${\cal L_C}(P) = \{\omega \in A^* | \gamma(s_0, \omega) \in g\}$ which is defined by the state transition system defined by the preconditions and effects of the primitive tasks. This language is regular.
\end{enumerate}

The key idea of our approach is to learn the DFA $\Sigma = (S,A,\gamma)$ corresponding to the regular language ${\cal L_C}(P)$, and modify the DFA to approximate the language ${\cal L}(P)$ with the DFA $\Sigma=(S,T,\gamma)$ and then infer $P$.

\section{The AMLSI Approach}\label{sec:backgroung}

In this section we will summarized the AMLSI approach on which HierAMLSI is based. For more detail see \cite{amlsi_ictai, amlsi_keps}.

AMLSI generates the set of observations $\Omega$ by using random walks to learn $\Sigma = (S, A, \gamma)$ and deduce $P = (L, A, S, s_0, G, \delta, \lambda)$. AMLSI assumes $L$, $A$, $S$, $s_0$ known and the observation function $\lambda$ possibly partial and noisy (a partial observation is a state where some propositions are missing and a noisy observation is a state where the truth value of a proposition is erroneous). No knowledge of the goal states $G$ is required. Once $\Sigma$ is learned, AMLSI has to deduce $\delta$ from the transition function $\gamma$. Concretely, $\delta$ can be represented as a STRIPS planning domain containing all the actions of the problem $P$ and by induction the classical PDDL operators.

The AMLSI algorithm consists of 4 steps: (1) generation of the observations, (2) learning the DFA corresponding to the observations, (3) induction of the PDDL operators from the learned DFA; (4) finally, refinement of these operators to deal with noisy and partial state observations:

\paragraph*{Step 1:} AMLSI generates a random walk by applying an action from the initial state of the problem. If the action is applicable in the current state the sequence of actions from the initial state is valid and is added to $I^{+}$, the set of positive samples. Otherwise the random walk is stopped and the sequence is added to $I_{-}$, the set negative samples.

\paragraph*{Step 2:} To learn the DFA $\Sigma = (S, A, \gamma)$ AMLSI uses a variant \cite{amlsi_ictai} of a classical regular grammar learning algorithm called RPNI \cite{rpni}. The learning is based on both $I^{+}$ and $I^{-}$.

\paragraph*{Step 3:} AMLSI begins by inducing the preconditions and effects of the actions. For the preconditions $prec(a)$ of action $a$, AMLSI computes the logical propositions that are in all the states preceding $a$ in $\Sigma$:
\begin{equation*}\label{eq:generation_precondition}
prec(a)=\cap_{s \in preset(a)}\lambda(s)
\end{equation*}
For the positive effects $add(a)$ of action $a$, AMLSI computes the logical propositions that are never in states before the execution of $a$, and always present after $a$ execution:
\begin{equation*}\label{eq:positive_generation_effect}
add(a) = \cap_{s \in preset(a)}\lambda(s) \setminus prec(a)
\end{equation*}
Dually,
\begin{equation*}\label{eq:negative_generation_effect}
del(a) = prec(a) \setminus \cap_{s \in postset(a)}\lambda(s)
\end{equation*}
Once preconditions and effects are induced, actions are lifted to PDDL operators based on OI-subsumption (subsumption under Object Identity) \cite{DBLP:journals/ml/EspositoSFF00}: first of all, constant symbols in preconditions and effects are substituted by variable symbols. Then, the less general preconditions and effects, i.e. preconditions and effects encoding as many propositions as possible, are computed as intersection sets. This generalization method allows to ensure that all the necessary preconditions, i.e. the preconditions allowing to differentiate the states where actions are applicable from states where they are not, to be rightfully coded in the corresponding operators.

\paragraph*{Step 4:} To deal with noisy and partial state observations, AMLSI starts by refining the operator effects to ensure that the generated operators allow to regenerate the induced DFA. To that end, AMLSI adds all the effects ensuring that each transition in the DFA are feasible. Then, AMLSI refines the preconditions of the operators. As in \cite{arms}, it makes the following assumptions: the negative effects of an operator must be used in its preconditions. Thus, for each negative effect of an operator, AMLSI adds the corresponding proposition in the preconditions. Since effect refinements depend on preconditions and precondition refinements depend on effects, AMLSI repeats these two refinements steps until convergence, i.e., no more precondition or effect is added. Finally, AMLSI performs a Tabu Search
to improve the PDDL operators independently of the induced DFA, on which operator generation is based. Once the Tabu Search reaches a local optimum, AMLSI repeats all the three refinement steps until convergence.

\section{The HierAMLSI approach}\label{sec:method}


HierAMLSI generates the set of observations $\Omega$ by using random walks to learn $\Sigma = (S, T, \gamma)$ and deduce $P = (L, C, A, S, M, s_0, w_I, g, \delta, \tau, \lambda, \sigma, \zeta)$. HierAMLSI makes the same assumptions than AMLSI: $L$, $A$, $S$, $s_0$ are known but also $C$, the decomposition function $\zeta$ and the observation function $\lambda$ possibly partial and noisy (a partial observation is a state where some propositions are missing and a noisy observation is a state where the truth value of a proposition is erroneous). No knowledge of the goal states $g$ is required. Once $\Sigma$ is learned, HierAMLSI has to deduce $\delta$ from the transition function $\gamma$ and $\sigma$ from the decomposition function $\zeta$.

Like AMLSI, HierAMLSI uses random walks to generate $\Omega$ and makes the same assumptions.

HierAMLSI has been devised to solve {\sf HTN} learning problems. In practice, it computes $P$ as HDDL domain and problem files \cite{hddl, hddl2}.

Regarding the training dataset, HierAMLSI uses random walks to generate $\Omega$. HierAMLSI makes the same assumptions than AMLSI. Once the DFA is learned, HierAMLSI generates the set of methods $\sigma$ and infers the action precondition, positive and negative effect functions in $\delta$ from the state transition function $\gamma$. Finally, the methods and operators of the HDDL domain file are induced from $\sigma$ and $\delta$.

The HierAMLSI approach consists of 4 steps:
\begin{enumerate}
    \item {\it Generation of the observations.} HierAMLSI produces a set of observations $\Omega$ by using a random walk. In section \ref{sec:method:data}, we will present how HierAMLSI is able to efficiently exploit these observations by taking into account not only the fact that some task are decomposable in certain states and their decomposition but also that others are not.
    \item {\it DFA Learning.} HierAMLSI learns a DFA approximating the language ${\cal L}(P)$ (see Section \ref{sec:method:grammar}).
    \item {\it {\sf HTN} Methods generation.} HierAMLSI generates from the DFA learned previously a set of {\sf HTN} Methods allowing to decompose all tasks observed in $\Omega$ (see Section \ref{sec:method:method_learning}).
    \item {\it Action model and {\sf HTN} Methods precondition learning.} Once {\sf HTN} Methods have been learned, HierAMLSI have to learn the action model, i.e. primitive tasks preconditions and effects and the {\sf HTN} Methods preconditions. To do this, HierAMLSI treat {\sf HTN} Methods as primitive tasks and learn an action model containing both methods and primitive tasks using the learning and refinement techniques proposed by the AMLSI approach (see Section \ref{sec:backgroung}).
\end{enumerate}

The rest of this Section will be illustrated using the IPC\footnote{\url{https://www.icaps-conference.org/competitions/}} Gripper domain. In this domain, a robot moves balls in different rooms using its two grippers $r$ and $l$. This domain contains three compound tasks: $goto(r_a)$ the robot goes into the room $r_a$, $move1ball(b_1~r_a)$ the robots move the ball $b_1$ in the room $r_a$ and $move2balls(b_1~b_2~r_a)$ the robot moves balls $b_1$ and $b_2$ in the room $r_a$.

\subsection{Observation Generation}
\label{sec:method:data}

The data generation process is similar to the generation method of the AMLSI algorithm \cite{amlsi_ictai}. To generate the observations in $\Omega$, HierAMLSI uses random walks by querying a blackbox. HierAMLSI chooses randomly a (primitive or compound) task $t$. If the task $t$ is decomposable, the blackbox returns the final decomposition $\phi$ containing only primitive task and this decomposition is added to the current primitive task sequence. This procedure is repeated until the selected task is not applicable to the current state. The applicable prefix of the primitive task sequence is then added to $I_{+}$, the set of positive samples, and the complete sequence, whose last task is not applicable, is added to $I_{-}$, the set of negative samples. Random walks are repeated until $I_{+}$ and $I_{-}$ achieve an arbitrary size.

\subsection{DFA Learning}
\label{sec:method:grammar}

\begin{figure*}[t]
    \centering
    \includegraphics[width=\linewidth]{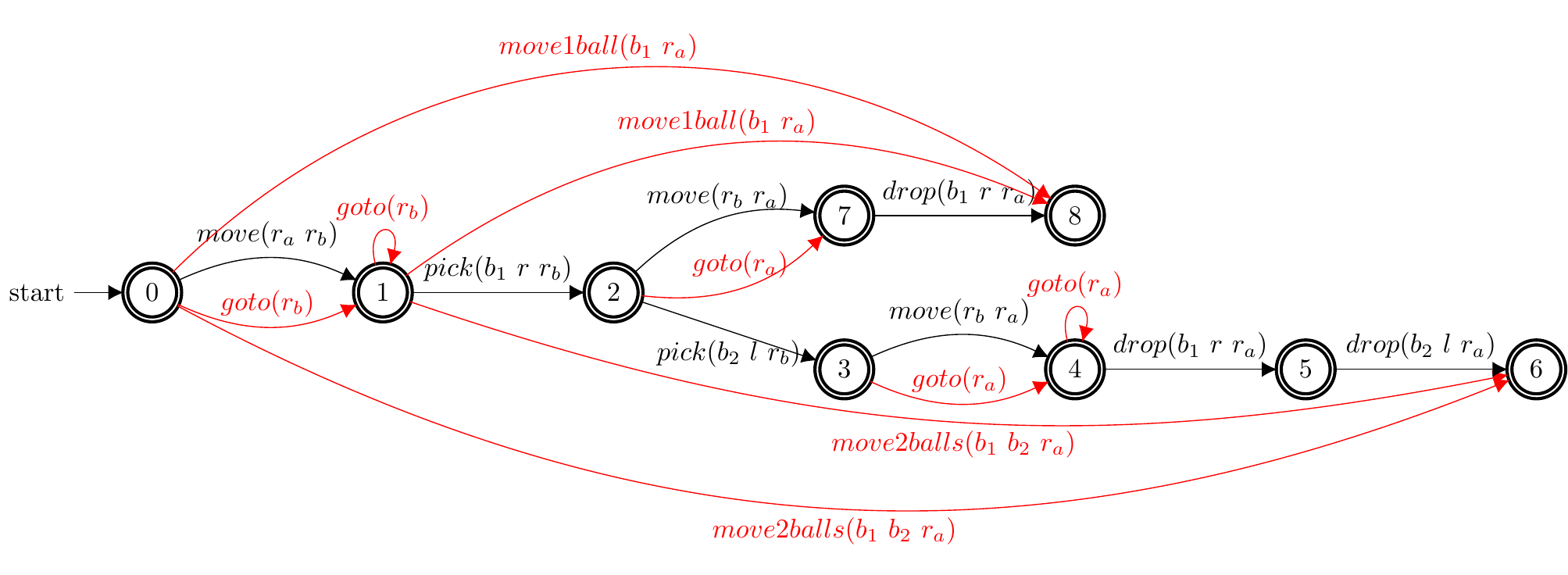}
    \caption{DFA learning steps  The Primitive Task DFA is the DFA containing only Primitive Task, i.e. black transition, and the Task DFA contains Compound Tasks, i.e. red transitions, in addition}
    \label{fig:dfa:learning}
\end{figure*}

As mentioned in Section \ref{sec:formal} the language ${\cal L}(P)$ is not necessary regular, then the purpose of this step is to learn a DFA approximating this language. More precisely, the DFA learning step is divided in 2 steps: (1) HierAMLSI learns a DFA corresponding to the language ${\cal L_C}(P)$ which is defined by the state transition system defined by the preconditions and effects of the primitive tasks and (2) HierAMLSI adds transitions to represent compound tasks in the DFA to allow to approximate the language ${\cal L}(P)$.

\paragraph*{Step 1: Primitive task DFA Learning}
HierAMLSI starts by using the DFA Learning algorithm proposed by \cite{amlsi_ictai} to learn the DFA containing only primitive tasks.

\paragraph*{Step 2: Task DFA Induction}
Once the Primitive Task DFA has been learned, HierAMLSI induces the Task DFA by adding compound task transition in the DFA, i.e. by adding transitions whose labels are compound task labels. For instance, suppose we have the compound task $move2balls(b_1~b_2~r_b)$ has been decomposed by primitive tasks $(move(r_a~r_b),$ $pick(b_1~r~r_b),$ $pick(b_2~l~r_b),$ $move(r_b~r_a),$ $drop(b_1~r~r_a),$ $drop(b_2~l~r_a))$ in node $0$ and reached the node $6$. Then we add the following transitions in the DFA $\gamma(move2balls(b_1~b_2~r_b), 0) \rightarrow 6$. Figure \ref{fig:dfa:learning} gives an example of Task DFA.

\begin{figure}[t]
    \centering
    \includegraphics{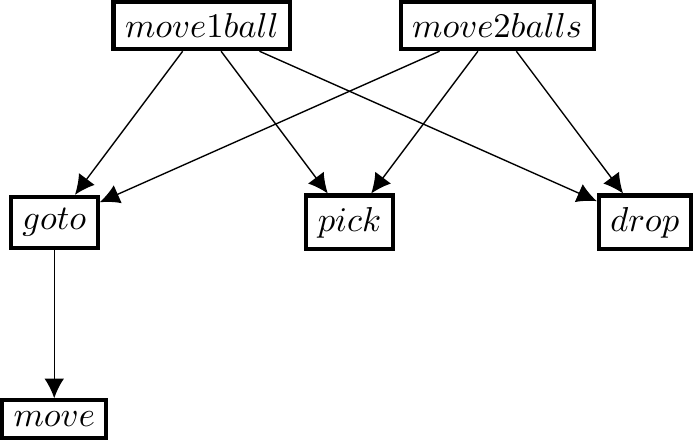}
    \caption{Tasks dependencies for the IPC Gripper domain}
    \label{fig:method:method_learning:dependency}
\end{figure}

\begin{algorithm}[t]
$Initialization(M)$\\
\For{$i = 1:|C|$}{
    \For{$c \in C$} {
        $M'[c] \leftarrow greedy(c, M, i)$\\
    }
    \For{$c \in C$} {
        \If{$|M'[c]| < |M[c]|$}{
            $M[c] \leftarrow M'[c]$\\
        }
    }
}
\Return{$M$}
\caption{Heuristic {\sf HTN} Methods learning}
\label{algo:heuristic}
\end{algorithm}

\begin{figure*}
\centering
\begin{subfigure}[t]{0.9\textwidth}
    \begin{equation*}
        \resizebox{\linewidth}{!}{$
        \begin{array}{ll}
            goto(?r) : & \phi_1=(), \phi_2=move(?r_1~?r)\\
            move2balls(?b_1~?b_2~?r) : & \phi_1= (pick(?b_1~?g_1~?r_1),pick(?b_1~?g_1~?r_1),move(?r_1~?r),drop(?b_1~?g_1~?r),drop(?b_1~?g_1~?r))\\
            & \phi_2=(move(?r~?r_1),pick(?b_1~?g_1~?r_1),pick(?b_1~?g_1~?r_1),move(?r_1~?r),drop(?b_1~?g_1~?r),drop(?b_1~?g_1~?r))\\
            move1ball(?b~?r) : & \phi_1 = (pick(?b~?g~?r_1),move(?r_1~?r),drop(?b~?g~?r))\\
            & \phi_2 = (move(?r_?~?r_1)),pick(?b~?g~?r_1),move(?r_1~?r),drop(?b~?g~?r))\\
        \end{array}
        $}
    \end{equation*}
    \caption{\textbf{Step 0:} Initialization with no compound task dependency}
    \label{fig:hieramlsi:methods:step0}
\end{subfigure}
\begin{subfigure}[t]{0.9\textwidth}
    \begin{equation*}
        \resizebox{\linewidth}{!}{$
        \begin{array}{ll}
            goto(?r) : & \phi_1=(), \phi_2=move(?r_1~?r)\\
            move2balls(?b_1~?b_2~?r) : & \phi_1=(goto(?r_1),pick(?b_1~?g_1~?r_1),pick(?b_1~?g_1~?r_1),goto(?r),drop(?b_1~?g_1~?r),drop(?b_1~?g_1~?r))\\
            move1ball(?b~?r) : & \phi_1 = (goto(?r_1),pick(?b~?g~?r_1),goto(?r),drop(?b~?g~?r))\}\\
        \end{array}
        $}
    \end{equation*}
    \caption{\textbf{Step 1:} Induction with 1 compound task dependence}
    \label{fig:hieramlsi:methods:step1}
\end{subfigure}
\begin{subfigure}[t]{0.9\textwidth}
    \begin{equation*}
        \resizebox{\linewidth}{!}{$
        \begin{array}{ll}
            goto(?r) : & \phi_1=(), \phi_2=move(?r_1~?r)\\
            move2balls(?b_1~?b_2~?r) : & \phi_1=(goto(?r_1),pick(?b_1~?g_1~?r_1),pick(?b_1~?g_1~?r_1),goto(?r),drop(?b_1~?g_1~?r),drop(?b_1~?g_1~?r))\\
            move1ball(?b~?r) : & \phi_1 = (goto(?r_1),pick(?b~?g~?r_1),goto(?r),drop(?b~?g~?r))\}\\
        \end{array}
        $}
    \end{equation*}
    \caption{\textbf{Step n:} Induction with $n$ compound task dependence}
    \label{fig:hieramlsi:methods:stepn}
\end{subfigure}
\caption{{\sf HTN} Methods learning example}
\label{fig:hieramlsi:methods}
\end{figure*}

\subsection{{\sf HTN} Methods Learning}
\label{sec:method:method_learning}

Once the Task DFA is induced HierAMLSI can directly extract {\sf HTN} Methods from the Task DFA. However, it is possible that a large number of {\sf HTN} Methods has been generated. For instance, let's take the Task DFA in Figure  \ref{fig:dfa:learning}. For the compound task $move1ball(?b~?r)$ HierAMLSI can generate several methods:
\begin{equation*}
\resizebox{\linewidth}{!}{$
\begin{array}{ll}
\phi_1 = & (move(?r~?r_1), pick(?b~?g~?r_1), move(?r_1~?r), drop(?b~?g~?r))\\
\phi_2 = & (pick(?b~?g~?r_1), move(?r_1~?r), drop(?b~?g~?r))\\
\phi_3 = & (goto(?r_1), pick(?b~?g~?r_1), move(?r_1~?r), drop(?b~?g~?r))\\
\phi_4 = & (goto(?r_1), pick(?b~?g~?r_1), goto(?r), drop(?b~?g~?r))\\
& \ldots \\
\end{array}
$}
\end{equation*}

Some of these decomposition are redundant. To facilitate proof reading we want a more compact description of the {\sf HTN} Methods. More precisely, we want minimizing the set of methods allowing to decompose observed compound tasks. Then, the {\sf HTN} Methods learning problem can be reduce to a variant of the set cover problem \cite{karp} which is NP-Complete.

The Greedy algorithm \cite{greedy} is a classical way to approximate the solution in a polynomial time. At each stage of the Greedy algorithm, HierAMLSI chooses the method that allows to decompose the largest number of tasks.

The main drawback of this approach is that it does not take into account the fact that there are dependencies between tasks. For instance, the optimal way to decompose the compound task $move1ball(?b~?r)$ is $\phi_4 = (goto(?r_1),$ $pick(?b~?g~?r_1),$ $goto(?r),$ $drop(?b~?g~?r))$ (see Figure \ref{fig:method:method_learning:dependency}). So the compound task $move1ball(?b~?r)$ depends of the compound task $goto(?r)$. So, as long as all methods for the compound task $goto(?r)$ have been generated, the Greedy algorithm will always prioritize $\phi_1$ and $\phi_2$ to $\phi_4$. Indeed, the decomposition $\phi_4$ can be added to the current solution of the Greedy algorithm if and only if all methods for the compound task $goto(?r)$ have been added.

\paragraph*{Heuristic Approach}

We propose a sound, complete and polynomial Heuristic approach (see Algorithm  \ref{algo:heuristic}) taking into account dependencies between tasks. Figure  \ref{fig:hieramlsi:methods} gives an example for the Gripper domain.

HierAMLSI starts by initializing the set of {\sf HTN} methods using the decomposition $\phi$ observed during the observation generation step (see Section  \ref{sec:method:data}). For each Compound Task we have therefore a set of {\sf HTN} Methods containing only Primitive Tasks and no Compound Task dependencies. For instance, for the Compound Task $move1ball(?b~?r)$ we have the two following decomposition:
\begin{equation*}
\resizebox{\linewidth}{!}{$
\begin{array}{ll}
    \phi_1 = & (pick(?b~?g~?r_1),move(?r_1~?r),drop(?b~?g~?r))\\
     \phi_2 = & (move(?r_?~?r_1)),pick(?b~?g~?r_1),move(?r_1~?r),drop(?b~?g~?r))\\
\end{array}$}
\end{equation*}
Then, at each iteration AMLSI use the greedy algorithm to compute a new set of {\sf HTN} Methods with an additional Compound Task dependency. Finally, if the new {\sf HTN} Methods set is smaller than the one learned in the previous iteration, then it is retained.
For instance, for the Gripper domain, suppose we have the two following decomposition for the Compound Task $goto(?r)$:
\begin{equation*}
\begin{array}{ll}
    \phi_1 = & ()\\
    \phi_2 = & (move(?r_1~?r)))\\
\end{array}
\end{equation*}
Then, the Greedy algorithm return only one decomposition for the Compound Task $move1ball(?b ?r)$: $(goto(?r_1)),$ $pick(?b~?g~?r_1),$ $goto(?r),$ $drop(?b~?g~?r))$.

\begin{lemme}
The Heuristic approach is sound and complete. The heuristic approach generates a set of {\sf HTN} Methods $M$ able to decompose all observed compound tasks in the dataset $\Omega$.
\end{lemme}

\begin{proof}
During the observation generation step (see Section \ref{sec:method:data}), for each generated Compound Task $t$, we have its final decomposition $\phi$. So at the initialization step of the Heuristic approach, there at least one method able to decompose each observed Task. The initialization is therefore sound and complete. Moreover the following steps of the Heuristic approach generates methods decomposing as many tasks as the previous steps, then the Heuristic approach is sound and complete.
\end{proof}

\begin{lemme}
The Heuristic approach is polynomial.
\end{lemme}

\begin{proof}
First of all, we have $\mathcal{O}(|I_{+}|)$\footnote{$|I_+|$ denote the number of primitive tasks in the positive sample} nodes in the DFA. Then, in the worst case, we have a possible {\sf HTN} Method for each node pair, then we have $\mathcal{O}(|I_{+}|^2)$ possible {\sf HTN} Methods in the DFA. Then, the complexity of the Greedy algorithm is $\mathcal{O}(|I_{+}|^3)$ in term of tested decomposition. Finally, according to the algorithm \ref{algo:heuristic}, the Greedy algorithm is repeated $|C|^2$ times. Finally, the complexity of the Heuristic approach is $\mathcal{O}(|C|^2.|I_{+}|^3)$.
\end{proof}

\begin{table}[!t]
\centering
\resizebox{\linewidth}{!}{
\begin{tabular}{|c||c|c|c|c|c|}
\hline
Domain & \# Primitive Task & \# Compound Task & \# Methods & \# Predicates\\
\hline\hline
Blocksworld & $4$ & $4$ & $8$ & $5$ \\ \hline
Gripper & $3$ & $3$ & $4$ & $4$ \\ \hline
Zenotravel & $4$ & $2$ & $5$ & $7$ \\ \hline
Transport & $3$ & $4$ & $6$ & $5$ \\ \hline
Childsnack & $6$ & $1$ & $2$ & $12$ \\ \hline
\end{tabular}
}
\caption{\label{tab:benchmarks} Benchmark domain characteristics. From left to right, the number of Primitive Tasks, the number of Compound Tasks, the number of Methods and the number of Predicates for each IPC domain.}
\end{table}

\begin{figure*}[!t]
    \centering
    \includegraphics[width=\linewidth]{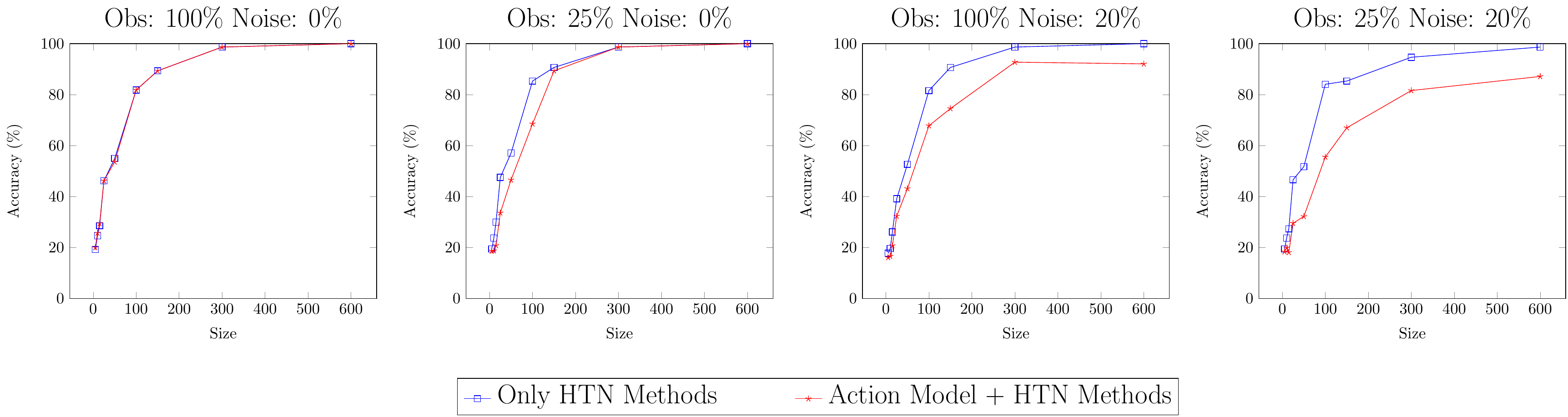}
    \caption{\label{fig:experiment:results} Performance of HierAMLSI when the training data set size increases in number of tasks in terms of Accuracy.}
\end{figure*}

\section{Exprimentation}\label{sec:experiments}

The purpose of this evaluation is to evaluate the performance of HierAMLSI though two variants: (1) we evaluate the performance of HierAMLSI when only {\sf HTN} Methods are learned, i.e. the action model is known and (2) we evaluate the performance of HierAMLSI when both {\sf HTN} Methods are learned and the action model is unknown. We use 4 experimental scenarios\footnote{Note that these are the experimental scenarios used to test AMLSI on which HierAMLSI is built }:
\begin{enumerate}
\item Complete intermediate observations (100\%) and no noise (0\%).
\item Complete intermediate observations (100\%) and high level of noise (20\%).
\item Partial intermediate observations (25\%) and no noise (0\%).
\item Partial intermediate observations (25\%) and high level of noise (20\%).
\end{enumerate}

\subsection{Experimental Setup}\label{sec:experiments:setup}

Our experiments are based on 5 HDDL \cite{hddl, hddl2} domains (see Table  \ref{tab:benchmarks}) from the IPC 2020 competition: Blocksworld, Childsnack, Transport, Zenotravel and Gripper.

HierAMLSI learns {\sf HTN} domains from one instance. To avoid performances being biased by the initial state, HierAMLSI is evaluated with different instances. Also, for each instance, to avoid performances being biased by the generated observations, experiments are repeated five times. All tests were performed on an Ubuntu 14.04 server with a multi-core Intel Xeon CPU E5-2630 clocked at $2.30$ GHz with 16GB of memory. PDDL4J library \cite{pddl4j} was used to generate the benchmark data.

\subsection{Evaluation Metrics}\label{sec:experiments:metrics}

HierAMLSI is evaluated using the {\em accuracy} \cite{rim} that measures the learned domain performance to solve new problems.

Formally, the accuracy $Acc = \frac{N}{N^{*}}$ is the ratio between $N$, the number of correctly solved problems with the learned domain, and $N^{*}$, the total number of problems to solve. In the rest of this section the accuracy is computed over 20 problems. The problems are solved with the TFD planner \cite{tfd} provided by the PDDL4J library. Plan validation is done with VAL, the IPC competition validation tool \cite{val}.

\subsection{Discussion}\label{sec:experiments:results}

Table  \ref{fig:experiment:results} shows the accuracy of HierAMLSI obtained on the 5 domains of our benchmarks when varying the training data set size. The size of the training set is indicated in number of tasks.

First of all, we observe that when HierAMLSI learns only the set of {\sf HTN} Methods, learned domains are generally optimal (Accuracy $= 100\%$) with $600$ tasks whatever the experimental scenario. Also, $100$ tasks are generally sufficient to learn accurate domains (Accuracy $> 50\%$). Then, when HierAMLSI learns both action model and {\sf HTN} Methods performances are similar when observations are noiseless. However, when observations are noisy, performances are downgraded for some domains: Blocksworld and Childsnack when observations are complete and Blocksworld, Transport and Childsnack when observations are partial. This is due to the fact that there are learning error in the action model learned. However, learned domains remain accurate when there are at least $300$ tasks in the training dataset.

To conclude, we have shown experimentally that HierAMLSI learns accurate domains. More precisely, when the action model is known, HierAMLSI generally learns optimal domains. Also the performances are downgraded when HierAMLSI has to learn the action model in addition to the set of {\sf HTN} methods, but the learned domains remain accurate. Performance degradation are due to the learning errors in the action model.

\begin{table*}[!t]
\centering
\resizebox{\textwidth}{!}{
\begin{tabular}{|c|c|c|c|c|c|c|}
\hline
\multirow{2}{*}{Algorithm}
&  \multicolumn{3}{|c|}{Input} &  \multicolumn{3}{|c|}{Output} \\ \cline{2-7}
& Input & Environment & Noise Level & Action Model & {\sf HTN} Methods & {\sf HTN} Methods Preconditions \\ \hline
\cite{xiaoa2019learning} & Plans & FO & $0\%$ & & $\bullet$ & \\ \hline
HTN-Maker & Plans & FO & $0\%$  & & $\bullet$ & \\ \hline
\cite{hogg2010learning} & Plans & FO & $0\%$  & & $\bullet$ & \\ \hline
HTN-Maker$^{\mbox{ND}}$ & Plans & FO & $0\%$ & & $\bullet$ & \\ \hline
\cite{pHTN} & Plans & FO & $0\%$  & & $\bullet$ & \\ \hline
\cite{learn_hierar} & Annoted Plans & PO & $0\%$ & & $\bullet$ & \\ \hline
LHTNDT & Annoted Plans & FO & $0\%$ & & $\bullet$ & \\ \hline
CAMEL & Plans & FO & $0\%$  & & $\bullet$ & $\bullet$  \\ \hline
HDL & Plans & FO & $0\%$  & & $\bullet$ & $\bullet$  \\ \hline
HTN-Learner & Decomposition Trees & PO & $0\%$  & $\bullet$ & $\bullet$ & $\bullet$ \\ \hline
HierAMLSI & Random Walks & PO & $20\%$  & $\bullet$ & $\bullet$ & $\bullet$ \\ \hline
\end{tabular}
}
\caption{\label{tab:related} State-of-the-art action model learning algorithms. From left to right: the kind of input data, the kind of environment: \textbf{F}ully \textbf{O}bservable or \textbf{P}artially \textbf{O}bservable, the maximum level of noise in observations, the kind of output}
\end{table*}

\begin{figure}
    \centering
    \begin{subfigure}[b]{0.45\textwidth}
         \centering
         \includegraphics[width=\textwidth]{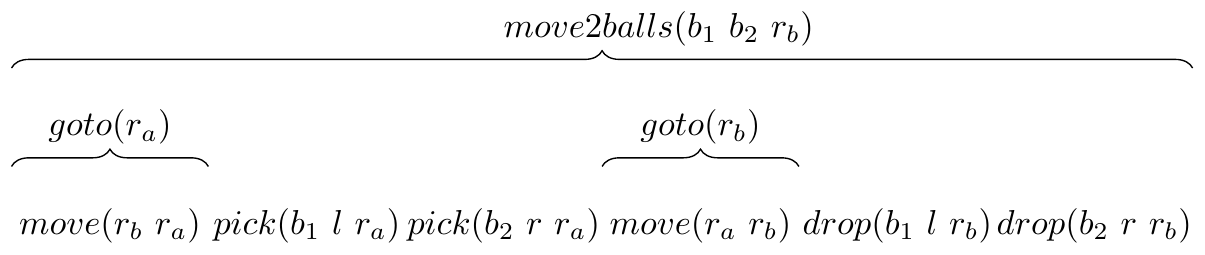}
         \caption{An annoted plan}
         \label{fig:input:annoted_plan}
     \end{subfigure}
     \begin{subfigure}[b]{0.45\textwidth}
         \centering
         \includegraphics[width=\textwidth]{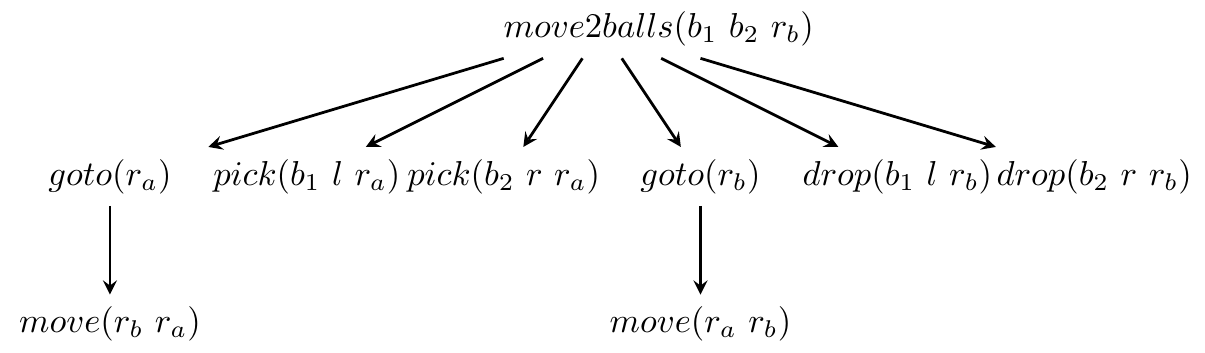}
         \caption{A decomposition tree}
         \label{fig:input:decomposition_tree}
     \end{subfigure}
    \caption{Input examples}
    \label{fig:input}
\end{figure}

\section{Related Works}\label{sec:related}

Many approaches have been proposed to learn HDDL domains. These approaches can be classified according to the input data of the learning process. The input data can be plan "traces" obtained by resolving a set of planning problems, annoted plans (see Figure  \ref{fig:input:annoted_plan}), decomposition tree (see Figure  \ref{fig:input:decomposition_tree}) or random walks. The input data can contain in addition to the tasks also states which can be fully observable (FO) or partially observable (PO), or noisy. Also, these approaches can be classified according to the output. The output can be the action model, the set of {\sf HTN} Methods and {\sf HTN} Methods preconditions. Table  \ref{tab:related} summarises these classifications.

A first group of approaches only learns the set of {\sf HTN} Methods. First of all, \cite{xiaoa2019learning} takes as input a set of plan traces and {\sf HTN} Methods and proposes an algorithm to update incomplete {\sf HTN} Methods by task insertions. Then HTN-Maker \cite{htn-maker} and HTN-Maker$^{\mbox{ND}}$ \cite{htn_maker_nd} takes as input plan trace generated from {\sf STRIPS} planner and annoted task provided by a domain expert. Then, \cite{hogg2010learning} proposed an algorithm based on reinforcement learning. Then, \cite{pHTN} proposed an algorithm taking as input only plan traces. This algorithm builds, from plan traces, a context free grammar (CFG) allowing to regenerate all plans. Then, methods are generated using CFG: one method for each production rule in the CFG. Then \cite{learn_hierar} and \cite{LHTNDT} proposed to learn {\sf HTN} Methods from annoted plan. Annoted plan are plan segmented with the different tasks solved. Figure  \ref{fig:input:annoted_plan} gives an example of annoted plan. However, obtaining these annotated examples is difficult and needs a lot of human effort.

A second group of approach learns {\sf HTN} Methods preconditions. First of all, the CAMEL algorithm \cite{camel} learns {\sf HTN} Methods and their the preconditions of {\sf HTN} Methods from observations of plan traces, using the version space algorithm. An annoted task is an triplet $(n, Pre, Eff)$ where $n$ is a task, $Pre$ is a set of proposition known as the preconditions and $Eff$ is a set of atoms known as the effects. These approach use annoted task to build incrementally {\sf HTN} Methods with preconditions. Then, the HDL algorithm \cite{DBLP:conf/aips/IlghamiNM06} takes as input plan traces. For each decomposition in plan traces, HDL checks if there exist a method responsible of this decomposition. If not, HDL creates a new method and initializes a new version space to capture its preconditions. Preconditions are learned in the same way as in the CAMEL algorithm.

Only HTN-Learner proposes to learn Action Model and {\sf HTN} Methods from decomposition trees. A decomposition tree is a tree corresponding to the decomposition of a method. Figure  \ref{fig:input:decomposition_tree} gives an example of decomposition tree.

\section{Conclusion}

In this paper we have presented HierAMLSI, a novel algorithm to learn HDDL domains. HierAMLSI is built on the AMLSI approach. HierAMLSI is composed of four steps. The first step consists, as AMLSI, in building two training sets of feasible and infeasible action sequences. In the second step, HierAMLSI induces a DFA. The third step is the generation of the {\sf HTN} Methods, and the last step learns {\sf HTN} Methods preconditions and the action model. Our experimental results show that HierAMLSI is able to learn accurately both action models and {\sf HTN} Methods from partial and noisy datasets.

Future works will focus on extending HierAMLSI in order to learn more expressive action model.

\section*{Acknowledgments}
This research is supported by the French National Research Agency under the "Investissements d’avenir” program (ANR-15-IDEX-02) on behalf of the Cross Disciplinary Program CIRCULAR.

\bibliography{biblio}

\end{document}